\pdfoutput=1
\DeclareUnicodeCharacter{202F}{\,}
\documentclass[11pt]{article}
\usepackage{paralist}
\usepackage{algpseudocode}
\usepackage{algorithm}
\usepackage{amsmath}
\usepackage{booktabs}        
\usepackage{multirow}        
\usepackage{array}           
\usepackage{makecell}        
\usepackage{amssymb}         
\usepackage{pifont}  
\usepackage[preprint]{acl}
\usepackage{url}
\usepackage{times}
\usepackage{latexsym}
\usepackage{enumitem}
\usepackage{tabularx}
\usepackage{multirow}
\usepackage{dblfloatfix}   

\usepackage{xcolor}
\definecolor{lightgray}{gray}{0.95}
\definecolor{medgray}{gray}{0.90}

\usepackage{tabularx}
\usepackage{multirow}
\usepackage[T1]{fontenc}
\usepackage{enumitem}

\usepackage[utf8]{inputenc}
\usepackage{ragged2e}
\usepackage{microtype}
\usepackage[export]{adjustbox}

\usepackage{listings}

\usepackage[many]{tcolorbox}

\usepackage[many]{tcolorbox}

\newtcolorbox{promptbox}[1][]{
  enhanced,
  colback=white,
  colframe=black!10,
  fonttitle=\bfseries\small,
  title=#1,
  coltitle=black,
  boxrule=0.3pt,
  arc=1pt,
  outer arc=1pt,
  left=6pt,
  right=6pt,
  top=4pt,
  bottom=4pt,
  sharp corners=south,
  before skip=10pt,
  after skip=10pt
}

\usepackage{inconsolata}
\usepackage{booktabs}
\usepackage{algorithm}
\usepackage{algpseudocode}
\usepackage{authblk}
\usepackage{graphicx}
\usepackage{caption} 
\usepackage{subcaption} 
\definecolor{darkblue}{RGB}{0, 0, 139}    
\definecolor{goldenyellow}{RGB}{255, 204, 0}  
\definecolor{lightblue}{RGB}{102,153,255}  
\definecolor{realpurple}{RGB}{153,51,255}
\definecolor{lightgray}{gray}{0.95}
\definecolor{medgray}{gray}{0.90}
\usepackage{tabularx}
\newcommand{\cmark}{\textcolor{green!70!black}{\ding{51}}} 
\newcommand{\xmark}{\textcolor{red}{\ding{55}}}  

\newcommand*\fullPDFht{945mm}   
\newcommand*\sliceHT{\textheight} 

\newcounter{nslice}
\title{OMS: On-the-fly, Multi-Objective, Self-Reflective \\Ad Keyword Generation via LLM Agent}

\author[1,2]{Bowen Chen\textsuperscript{*}}
\author[1]{Zhao Wang\textsuperscript{*}}
\author[1]{Shingo Takamatsu}

\affil[1]{Sony Group Corporation, Japan}
\affil[2]{The University of Tokyo, Japan}

\begin{document}
\maketitle

\begingroup
\renewcommand\thefootnote{}\footnotetext{\noindent \textsuperscript{*} indicates equal contribution. Bowen Chen completed this work during his internship at Sony. Corresponding author: Zhao.Wang@sony.com}
\endgroup

\begin{abstract}

Keyword decision in Sponsored Search Advertising is critical to the success of ad campaigns. While LLM-based methods offer automated keyword generation, they face three major limitations: reliance on large-scale query–keyword pair data, lack of online multi-objective performance monitoring and optimization, and weak quality control in keyword selection. These issues hinder the agentic use of LLMs in fully automating keyword decisions by monitoring and reasoning over key performance indicators such as impressions, clicks, conversions, and CTA effectiveness. To overcome these challenges, we propose \textbf{OMS}, a keyword generation framework that is \textit{On-the-fly} (requires no training data, monitors online performance, and adapts accordingly), \textit{Multi-objective} (employs agentic reasoning to optimize keywords based on multiple performance metrics), and \textit{Self-reflective} (agentically evaluates keyword quality). Experiments on benchmarks and real-world ad campaigns show that OMS outperforms existing methods;  Ablation and human evaluations confirm the effectiveness of each component and the quality of generated keywords. 


\begin{table*}[t]
    \centering
    \begin{minipage}[b]{0.5\textwidth}
        \centering
        \includegraphics[width=1\linewidth]{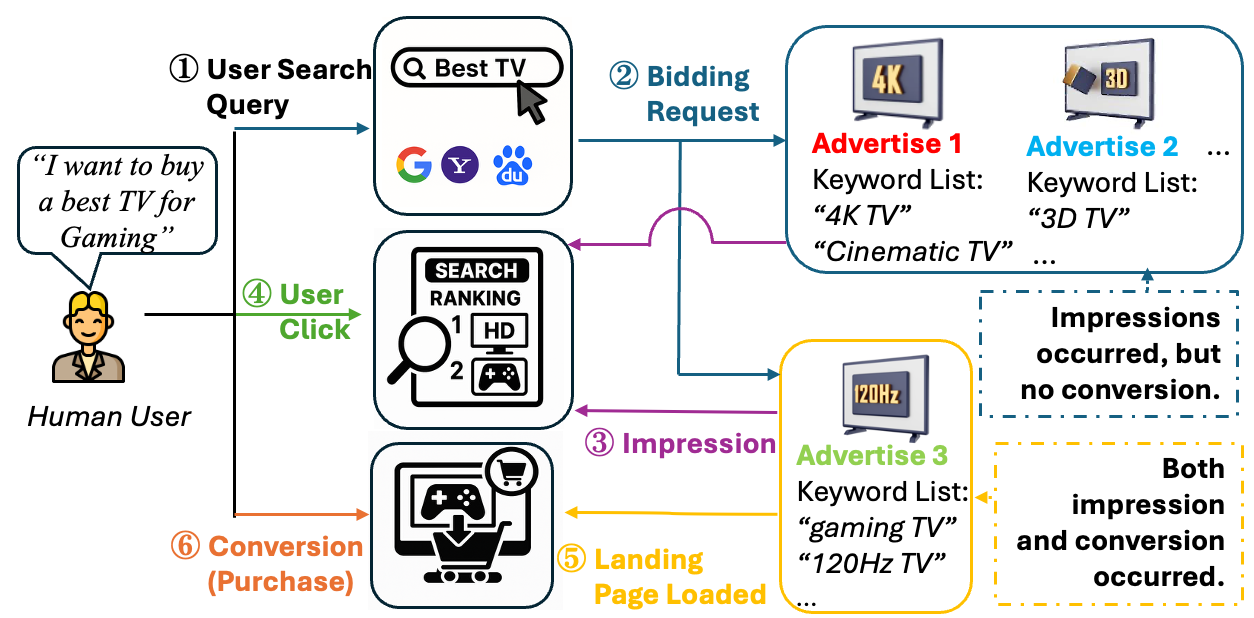}
        \captionof{figure}{An illustrative example: among the three advertisers (\textcolor{darkblue}{blue blocks} and \textcolor{goldenyellow}{yellow block}) that received impressions, only one (\textcolor{goldenyellow}{yellow block}) achieved a conversion.}
        \label{fig:index}
    \end{minipage}\hfill
    \begin{minipage}[b]{0.48\textwidth}
        \centering
        \scriptsize
        \setlength{\tabcolsep}{3pt}
         \resizebox{\textwidth}{!}{
            \begin{tabular}{lcccc}
                \toprule
                \multirow{2}{*}{\textbf{Method}} & \multicolumn{4}{c}{\textbf{Capabilities}} \\
                \cmidrule(lr){2-5}
                & \textbf{Free to } & \textbf{Real-Time} & \textbf{Multi-Objective}  & \textbf{Self-Reflective} \\
                & \textbf{Train} & \textbf{Monitoring} & \textbf{Optimization}  &\textbf{Quality Control} \\
                \midrule
                WIKG & \xmark & \xmark & \xmark &  \xmark \\
                GAN-based & \textbf{\xmark} & \textbf{\xmark} & \textbf{\xmark} & \textbf{\xmark} \\
                LSTM-based & \textbf{\xmark} & \textbf{\xmark} & \textbf{\xmark}  & \textbf{\xmark} \\
                LKG & \xmark & \xmark & \cmark  & \xmark \\
                OKG & \cmark & \cmark & \xmark & \xmark \\
                \textbf{Ours} & \textbf{\cmark} & \textbf{\cmark} & \textbf{\cmark}  & \textbf{\cmark} \\
                \bottomrule
            \end{tabular}
        } 
        \captionof{table}{Capability comparison between the proposed OMS method and conventional methods for the SSA keyword generation.}
        \label{tab:method_comparison}
    \end{minipage}
\end{table*}
\end{abstract}

\section{Introduction}

In Sponsored Search Advertising (SSA)~\cite{Fain2006SponsoredSA}, advertisers participate in bidding when their keyword lists broadly match the user's query. 
Figure~\ref{fig:index} shows how ad keywords affect the SSA process.
When a user enters a search query, the platform runs a real-time auction. Advertisers’ keyword lists decide whether their ads appear, where they are placed, and how likely users are to click or convert.
This highlights that while impression opportunities may be broadly shared, conversion likelihood heavily depends on how well an advertiser’s keyword list ~\cite{10.1145/1526709.1526873, 10.1145/1341531.1341564} captures the user’s intent, making keyword setting a critical factor in SSA.

Even today, automating keyword decisions beyond SSA platforms like Google, Bing \cite{google_keyword_planner, microsoft_keyword_planner} remains challenging, as only these platforms have access to large-scale query–keyword data.
Companies typically rely on keyword suggestion tools of SSA platforms or specialized ad agencies. However, even for SSA platforms and agencies, selecting effective keywords still requires ongoing effort—monitoring performance, updating keywords, and investing time and budget~\cite{wang-etal-2025-okg} to keep up with changing market trends and improve campaign outcomes.


In academic research, early methods used generative models such as GANs \cite{10.1145/3219819.3219850} and LSTMs \cite{lian2019endtoendgenerativeretrievalmethod} to generate keywords under this research direction. Recently, the emergence of large language models (LLMs) has helped overcome data scarcity, enabling advertisers to generate their own keywords. In this line of research, LKG \cite{10.1145/3589335.3651943} fine-tuned an LLM with constrained decoding to generate SSA keywords from real user queries. Meanwhile, OKG~\cite{wang-etal-2025-okg} introduces a keyword generation agent that leverages the reasoning ability of a ReAct-style LLM agent~\cite{yao2023react} to incorporate click feedback into the generation process, whose agentic reasoning helps identify underperforming keywords, missed intents, and emerging trends.

Despite these advancements, existing SSA keyword generation methods still face several limitations, as summarized in Table~\ref{tab:method_comparison}. Traditional approaches such as WIKG~\cite{8822966}, GAN-based~\cite{10.1145/3219819.3219850}, and LSTM-based~\cite{lian2019endtoendgenerativeretrievalmethod} methods are tightly dependent on large-scale data to train the models, and lack real-time performance monitoring and multi-objective optimization abilities. 
LKG~\cite{10.1145/3589335.3651943}, while leveraging LLMs trained on a static keyword dataset, lacks support for real-time feedback or reasoning during keyword generation, limiting its ability to adapt to rapid changes in user behavior.
OKG~\cite{wang-etal-2025-okg} has two major limitations. First, its decision-making is coarse-grained—it relies on predefined categories and aggregated click statistics, without jointly considering other important metrics such as conversions, CTA rates, and more. Second, its agentic process is static, simply following prompt-then-generate flow, which often results in low-quality keyword outputs.

We propose OMS, an On-the-fly, Multi-Objective and Self-Reflective ad keyword generation framework to address the above limitations and realize the functionalities outlined in Table~\ref{tab:method_comparison}. 
Specifically, our contributions are three-fold:

\begin{itemize}[left=0pt, itemsep=1pt]
\item \textbf{We propose an agentic clustering-ranking module} that monitors and clusters the currently active keywords, computes multi-objective scores, and ranks keywords both intra- and inter- clusters to identify high-impact keywords.

\item \textbf{We introduce a multi-turn generation-reflection module.} New keywords are generated through agentic reasoning and iteratively refined based on self-reflective feedback.

\item \textbf{Extensive Experimental Results} demonstrate that OMS outperforms multiple baselines on both established benchmarks and real-world online ad campaigns. Ablation studies highlight the effectiveness of each component, and human preference evaluations confirm OMS’s superiority across multiple metrics.
\end{itemize}

\section{Related Works}


\subsection{LLM Agents in Business Applications}

LLM agents have demonstrated their potential in business areas, including financial question answering \cite{Fatemi_2024}, financial sentiment analysis \cite{Xing_2025}, customer support \cite{zhong2023agentverse}, outperforming conventional methods.

While LLM agents have shown strong performance across many business domains, most existing LLM-based approaches in advertising have not yet embraced the agentic paradigm. 
Recent studies have explored the use of LLMs for ad text generation \cite{mita-etal-2024-striking, wang2025talkhier} and ad image creation \cite{chen2025ctr}, but these methods typically either do not leverage the full reasoning capabilities of LLM agents \cite{chen2025ctr}, or lack adaptive generation for real-time business performance \cite{mita-etal-2024-striking, article}.
As a result, such approaches remain confined to static, offline settings and struggle to cope with the dynamic, fast-changing nature of real-world advertising campaigns \cite{wang-etal-2025-okg}, where factors such as brand reputation, product quality, and market trends evolve constantly.

\subsection{Keyword Generation in SSA}
Early research utilizes knowledge graphs like Wikipedia \cite{8822966} to generate seed keywords and expand those keywords by statistical information or concept hierarchy-based expansions \cite{4063677, 10.1145/1526709.1526873, 10.1145/1341531.1341564}.
Later studies used GAN and LSTM models to generate keywords from user queries \cite{10.1145/3219819.3219850,lian2019endtoendgenerativeretrievalmethod}, but these relied heavily on proprietary data, restricting advertisers' ability to customize keyword strategies.
Recently, LLMs with their strong text generation ability have provided advertisers with opportunities to generate keywords.
\citet{10.1145/3589335.3651943} fine-tuned LLMs using query-keyword with click data and used beam-search to generate keywords.
OKG \cite{wang-etal-2025-okg} further advanced this field by integrating product information and real-time click performance data, enabling partial real-time adaptation.
However, as detailed in Section 1, the click aggregation based optimization and prompt-then-generate static agentic flow often results in low-quality keyword outputs.

\begin{figure*}[t] 
\centering
\includegraphics[width=0.99\linewidth, trim=0 150 390 140, clip]{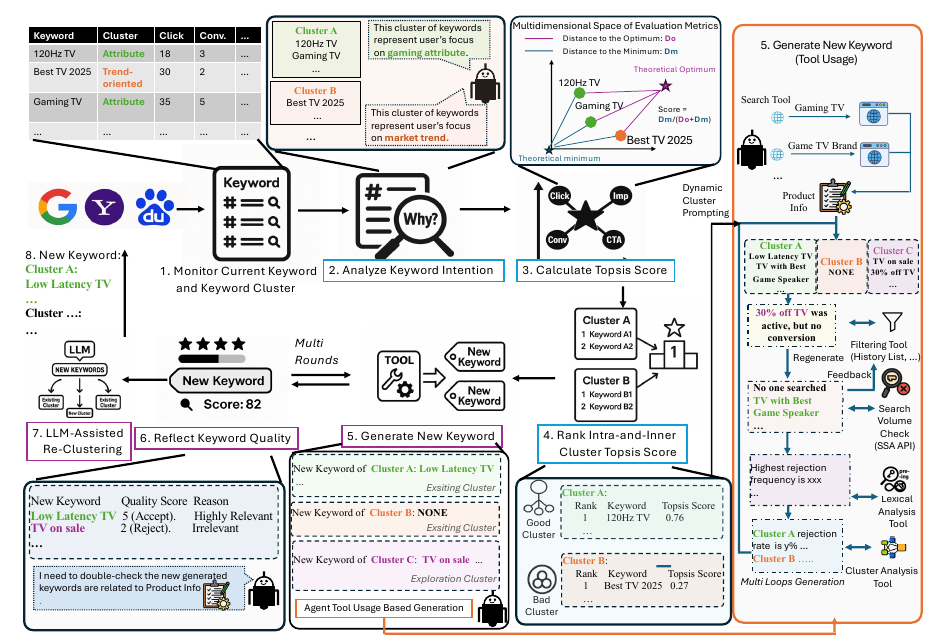}
\caption{Overview of the OMS workflow. The process consists of two main modules: \textcolor{lightblue}{\textbf{Agentic Clustering-Ranking}} (Steps 2–4) and \textcolor{realpurple}{\textbf{Multi-Turn Generation-Reflection}} (Steps 5–7). The system monitors keyword performance (Step 1), ranks keywords based on multi-objective analysis (Steps 2–4), generates new candidates, iteratively refines them and re-clusters the new generated keywords (Steps 5–7) before deployment (Steps 8).}
\label{fig:main}
\end{figure*}

\section{Problem Statement}
Given a product \( R \) advertised on a SSA platform with product information \( r \), at time step \( t \) during the ad compagin period, we have the historical keywords \( K_{\leq t} = \{k_1, k_2, \dots, k_I\} \) up to time \( t \), where \( I \) is the total number of historical keywords. 
Each keyword \( k_i \) (used at any time step \( \leq t \)) has a performance vector \( \mathbf{p}_i^{(\leq t)} = \{p_i^{1, (\leq t)}, p_i^{2, (\leq t)}, \dots, p_i^{m, (\leq t)}\} \), consisting of some relevant metrics such as Click, Conversion and etc.






Given \( K_{\leq t} \) and the associated performance vectors \( \{\mathbf{p}_i^{(\leq t)}\}_{i=1}^I \), a new keyword set \( K_t \) is generated at each time step \( t \) by a generation algorithm \( g \):

\begin{equation}
K_t = g(r, K_{\leq t}, \{\mathbf{p}_i^{(\leq t)}\}_{i=1}^I)
\end{equation}

The objective over the entire advertising campaign period \( T \) is to maximize the cumulative performance of all selected keywords:

\begin{equation}
\begin{aligned}
\max_{K_1, \dots, K_T} \quad & \sum_{t=1}^{T} \sum_{k_i \in K_t} \mathcal{W}^\top \mathbf{p}_i^{(t)} \\
\text{s.t.} \quad & \sum_{t=1}^{T} \sum_{k_i \in K_t} \text{Cost}(k_i^{(t)}) \leq B, \\
& |K_t| \leq N, \quad \forall t \in [1, T]
\end{aligned}
\label{formu::objective}
\end{equation}

\noindent \( \mathcal{W} \in \mathbb{R}^m \) is the weight vector over performance metrics; \( B \) is the total budget available for the entire campaign.
\footnote{Advertisers set the budget.
SSA restricts the number of deployable keywords. 
The budget usage, like bidding strategy for keywords, is managed automatically by the SSA platform.}
\( N \) is the maximum number of keywords that can be generated at each time step \( t \).

\section{Methodology}

Figure~\ref{fig:main} shows the overall OMS workflow process. 
The first module, \textbf{Agentic Clustering-Ranking} (Steps 2–4 in Figure~\ref{fig:main}), analyzes keyword intent, calculates multi-objective TOPSIS scores, and ranks keywords within and across clusters (Section~\ref{sec:clustering_ranking}).
The second module, \textbf{Multi-Turn Generation-Reflection} (Steps 5–7), dynamically formulates prompts, generates new keywords using external tools, iteratively refines them based on quality feedback and alysis, and re-clusters new generated keywords (Section~\ref{sec:generation_reflection}).
\footnote{All the prompts we used in this section can be found in Appendix~\ref{app::prompt} due to space limitations.}

\subsection{Agentic Clustering-Ranking Module}
\label{sec:clustering_ranking}


\subsubsection{Keyword Intent Analysis}

At each time step \( t \), we are given the current set of active (being set up in SSA) keywords \( K_{\leq t} \), along with their corresponding cluster assignments
$\mathcal{C}_{\leq t} = \{C_1, C_2, \dots, C_J\}$.
$\mathcal{C}_{\leq t}$ is calculated from the previous step \( t-1 \) by using eq.\ref{formu::recluter}.
Each cluster \( C_j \) contains a semantically coherent subset of keywords. 
Given the product description \( r \) and keywords in a cluster \( C_j \), OMS agent outputs an intent summary \( I_{C_j} \) this cluster:
\begin{equation}
\label{eq::intent}
I_{C_j} = \texttt{LLM\_Intent}(r, C_j), \quad \forall C_j \in \mathcal{C}_{\leq t}
\end{equation}

\noindent Here, \(\texttt{LLM\_Intent}\) is a prompt template used to acquire keyword intent. The resulting \( I_{C_j} \) provides an explanation of the likely search or marketing intent behind each cluster, such as emphasizing product features (e.g., ``high resolution'') or addressing consumer concerns (e.g., ``cost performance''). This intent summary enables our agent to reason about the current keyword setup and identify potential areas that may benefit from keyword expansion.

\subsubsection{TOPSIS Score Calculation and Ranking}

To prioritize keywords based on multiple performance metrics, we compute a TOPSIS score \( S(k_i) \in [0,1] \) for each active keyword \( k_i \in K_{\leq t} \) \cite{topsis}. Each performance vector \( \mathbf{p}_i^{(\leq t)} = \{p_i^1, \dots, p_i^m\} \) is first normalized using min-max scaling, where positive metrics (e.g., Clicks, Conversions) and negative metrics (e.g., Cost) are handled accordingly. Let \( v_{id} \in [0,1] \) denote the normalized value of keyword \( k_i \) for performance metric \( d \in \{1, \dots, m\} \). The weight vector \( \mathcal{W} = \{w_1, w_2, \dots, w_m\} \) reflects the importance of each metric and can be predefined as hyper-parameter or automatically adapted.

For each keyword with its normalized performance vector \( \{v_{i}\} \), we calculate its TOPSIS score \cite{hwang1981topsis}  based on its relative distance to the ideal (all-ones) and anti-ideal (all-zeros) coordinates:

\begin{equation}
S(k_i) = \frac{
    \sqrt{\sum_{d=1}^{m} w_d v_{id}^2}
}{
    \sqrt{\sum_{d=1}^{m} w_d (v_{id} - 1)^2} + \sqrt{\sum_{d=1}^{m} w_d v_{id}^2}
}
\end{equation}

\paragraph{Entropy-based Weighting.}
If weights \( \mathcal{W} \) are not predefined, we compute them using entropy. Let \( s_{id} = \dfrac{v_{id}}{\sum_{i=1}^{I} v_{id}} \) be the normalized contribution of keyword \( k_i \) to metric \( d \). The entropy-based weight \( w_d \) for each metric is then computed as:

\begin{equation}
w_d = 
\frac{ 1 - \dfrac{1}{\ln I} \sum_{i=1}^{I} s_{id} \ln s_{id} }
     { \sum_{d'=1}^{m} \left[ 1 - \dfrac{1}{\ln I} \sum_{i=1}^{I} s_{id'} \ln s_{id'} \right] }
\end{equation}

\noindent
Metrics with higher variance (lower entropy) receive larger weights, as they contain more potential space for keyword performance improvement.

\paragraph{Intra- and Inter-Cluster Ranking.}
For each cluster \( C_j \in \mathcal{C}_{\leq t} \), we perform \textit{intra-cluster ranking} by sorting the keywords \( k_i \in C_j \) in descending order of their TOPSIS scores \( S(k_i) \). This reveals the most and least effective keywords within each intent group and supports keyword-level refinement. 

We also compute a cluster-level score to enable \textit{inter-cluster ranking} across different keyword groups: \( S(C_j) = \frac{1}{|C_j|} \sum_{k_i \in C_j} S(k_i) \). Clusters are then ranked by their \( S(C_j) \) scores to prioritize high-performing groups for expansion and flag underperforming ones for improvement.

\begin{table*}[t]
\centering
\scriptsize
\renewcommand{\arraystretch}{1.15}
\setlength{\tabcolsep}{5pt}
\begin{tabular}{@{}p{3.3cm}p{12.2cm}@{}}
\toprule
\textbf{Tool} & \textbf{Description} \\ \midrule
\textbf{Search Tool} & Uses SerpAPI\footnote{\url{https://serpapi.com/}} to retrieve product information via Google queries until enough data is deemed by the Agent to be gathered. \\

\textbf{Reject Reflection} & Analyzes poor-performing keywords and provides feedback to the Keyword Generator. \\

\textbf{Keyword Filter Tool} & Flags validated low-performance or deployed keywords and asks the generator to replace them. \\


\textbf{Search Volume Validator} & Calling API to check if keywords meet search volume thresholds to avoid automatic rejection by the SSA platform. \\

\textbf{Keyword Lexical Analysis} & Detects common lexical patterns in SSA rejected keywords and provides an analysis report to avoid similar keywords. \\

\textbf{Category Analysis} & Identifies clusters with high rejection rates and prompts the generator to replace keywords of the entire cluster. \\ \bottomrule
\end{tabular}
\vspace{-1em}
\caption{Tools integrated into the OMS keyword generation framework.}
\label{tab:OMKG_tools}
\end{table*}


\subsection{Multi-Turn Generation-Reflection Module}
\label{sec:generation_reflection}


\paragraph{Dynamic Cluster Expansion for Many-Shot Prompting.}

To adaptively construct keyword generation prompts, we adaptively formulate a ranking-aware prompt \( \texttt{LLM\_Rank}(C_j) \) based on the cluster-level TOPSIS score and a predefined threshold \( \lambda \). 
The prompt for each cluster \( C_j \) is constructed as the following:

\begin{align}
\small
\label{eq::rank}
\texttt{LLM\_Rank}(C_j) = \begin{cases}
\left[ I_{C_j}; \{(k_i, S(k_i)) \mid k_i \in C_j \} \right], \\
\quad\quad\text{if } S(C_j) \ge \lambda \\
\left[ I_{C_j}; \{(k_i, S(k_i)) \mid k_i \in \texttt{TB}(C_j) \} \right], \\
\quad\quad\text{otherwise}
\end{cases}
\end{align}

\noindent
where \( I_{C_j} \) is the intent summary of cluster \( C_j \) from eq.\ref{eq::intent}, and \( S(k_i) \) is the TOPSIS score of keyword \( k_i \). 
The function \( \texttt{TB}(C_j) = \left\{ \arg\max_{k_i \in C_j} S(k_i) \right\} \cup \left\{ \arg\min_{k_i \in C_j} S(k_i) \right\} \) extracts both the highest- and lowest-ranked keywords for a bad performance cluster while all keywords information are provided in a good performance cluster.

This prompt expands high-performing clusters to contribute rich, many-shot context, while summarizing low-performance clusters with contrastive examples to guide generation direction.
\paragraph{Tool-Return-Value-Guided Generation Flow.}

Using the adaptive prompt above, OMS adopts a \textit{tool-return value-guided} generation flow. Rather than letting the LLM autonomously decide which tool to invoke.
Tools like the search volume checker, lexical analyzer, or filter (see Table~\ref{tab:OMKG_tools}) return explicit signals that guide the next step. 

For instance, if some keywords are flagged for low historical conversion, this tool will ask to re-generate them. 
If the SSA API flags low search volume keywords, it informs the OMS to call the lexical tool to analyze their lexical features. 
This multi-turn process (Figure~\ref{fig:main}, Step 5) improves interpretability and control. \footnote{Details of each tool are in the Appendix Sec \ref{tools}.}

\paragraph{Keyword Quality Reflection.}

After generation, OMS performs a reflection step to evaluate the quality of each keyword \( k_i \in K_t \), using the product information \( r \) and peer set \( K_t \). 
This process is formulated as following:

\begin{equation}
\texttt{LLM\_Reflect}(k_i) = \texttt{LLM}(k_i, K_t, r), \quad \forall k_i \in K_t
\label{eq::reflect}
\end{equation}

\noindent
The $\texttt{LLM\_Reflect}$ returns feedback (for example, “redundant”, “irrelevant to product”), and suggestions regarding whether to keep or regenerate certain keywords. 
This cycle continues until the agent internally deems the keyword set \( K_t \) complete.

\paragraph{LLM-Assisted Re-Clustering.}

Finally, OMS re-clusters the full keyword set \( K_{\leq t} \cup K_t \) using Affinity Propagation~\cite{doi:10.1126/science.1136800}, resulting in updated clusters \( \mathcal{C}_t = \{C_1, C_2, \dots, C_J\} \). 
For each new keyword \( k_i \in K_t \), we identify its top-3 closest clusters in embedding space, denoted \( \mathcal{C}_{k_i}^{\text{top}} \).
A final assignment is determined by:

\begin{equation}
C_{k_i} = \texttt{LLM\_Assign}(k_i, \mathcal{C}_{k_i}^{\text{top}})
\label{formu::recluter}
\end{equation}

\noindent
\( \texttt{LLM\_Assign} \) selects the best-fit cluster or creates a new one to ensure semantic coherence for new keywords. 
This step is meaningful because keywords are often short and mutually similar, so relying solely on embedding distance can reduce cluster quality, degrading the performance at the next step by introducing semantically unrelated keywords for the intent analysis for clusters.

\section{Experiments}
\noindent
\textbf{LLM Backbone and Baselines.}
We use GPT-4o \citep{openai2024gpt4ocard} as the backbone of OMS with temperature set to 0. Only in \texttt{LLM\_Reflect}, we use o3 as the reflection LLM to leverage its stronger reasoning ability. In our experiments, we compare the OMS with the following baselines:

\noindent \textbf{Google KW} \cite{google2024keywordplanner}: The built-in keyword planner tool provided by Google Ads.

\noindent \textbf{GPT Series} \cite{openai2024gpt4ocard}: GPT-4o, GPT-4.1, and o3 models prompted with the same product information and instruction to generate keywords.

\noindent \textbf{OKG} \cite{wang-etal-2025-okg}: We include the original OKG and \textit{Many-Shot OKG}, which augments OKG with examples, e.g, prompting each keyword with its performances on each metric.

\noindent
\textbf{Metrics.} We use following metrics:

\noindent \textbf{ROUGE-1 \cite{lin-2004-rouge}} compares the words overlap between generated keywords with the product information to compare the lexical coverage.

\noindent \textbf{BERTScore \cite{zhang2020bertscoreevaluatingtextgeneration}} considers both semantic and lexical similarity between generated keywords and the provided product information.

\noindent \textbf{Click, Search Volume, Cost per Click (CpC),} and \textbf{Competitor Score} are four keyword performance metrics: Click and Search Volume represent the number of campaign clicks and user searches for keywords, CpC indicates keyword efficiency, and Competitor Score (judged by the SSA platform) reflects keywords' competitiveness.

\begin{figure*}[htbp] 
\centering
\includegraphics[width=0.99\textwidth]{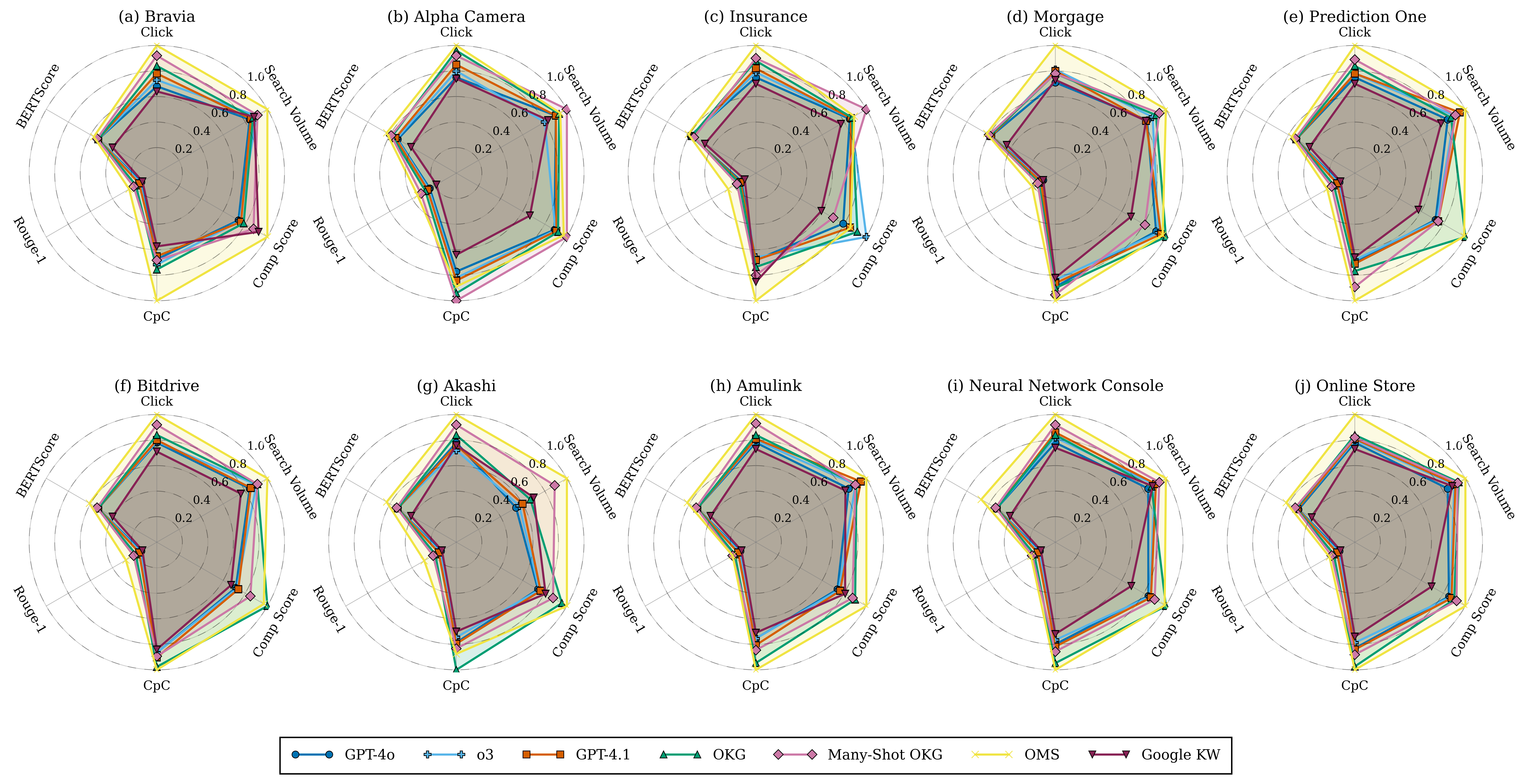} 
\caption{Performance Results on Benchmark Dataset. CpC is presented reversely, meaning a higher value means higher performance. Comp Score means the competitor score. The Rouge-1 and BERTScore are presented with raw values as their original values are between 0-1. Higher values indicate higher performance for all metrics.}
\label{fig::vis}
\end{figure*}

\begin{figure*}[t]   
  \centering
  \includegraphics[width=1\textwidth]{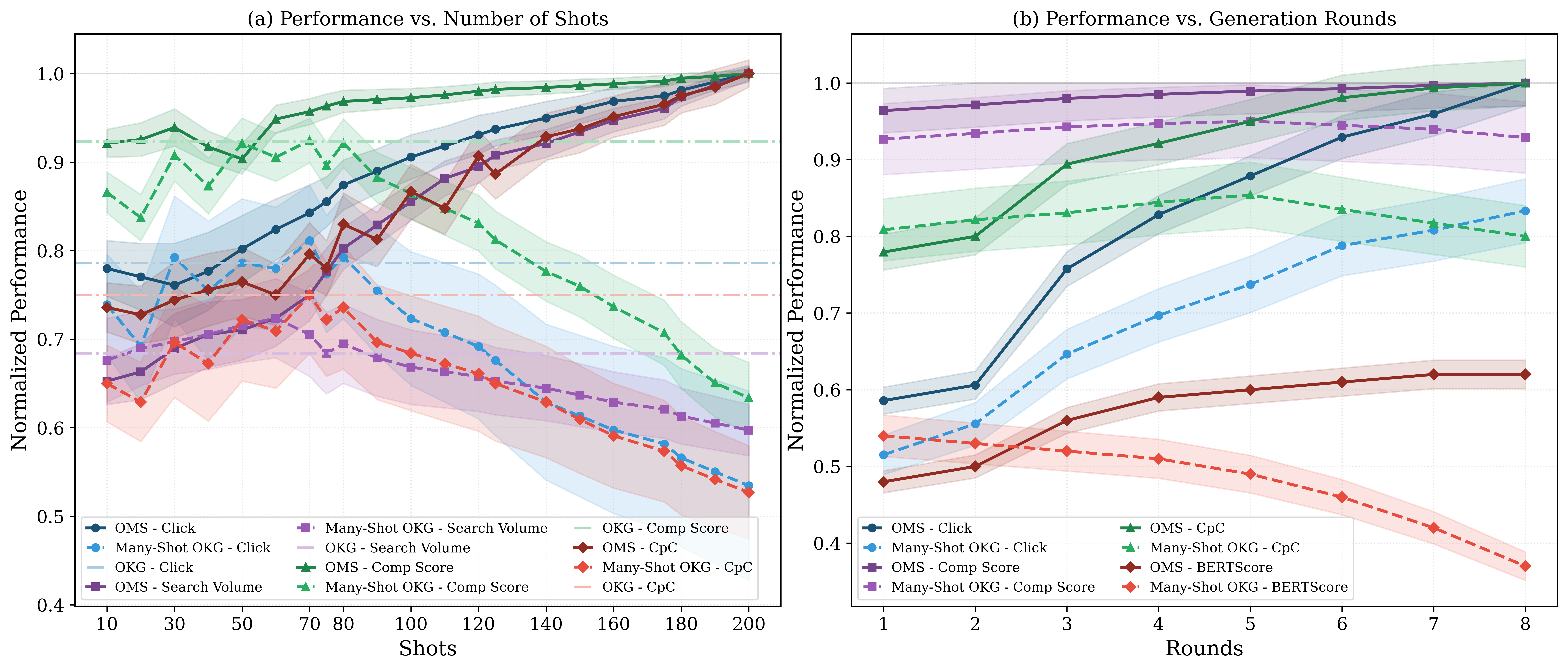}
    \caption{Performance change with Number of Shots and Generation Rounds (Under 80 examples).}
    \label{fig:shots}
\end{figure*}

\paragraph{Offline Benchmark Dataset.}
We use the benchmark released by OKG \cite{wang-etal-2025-okg}.
It collects keyword performance in the real world for 10 products with 1,000 keywords for each product.

\paragraph{Online A/B Test.}
We deploy OMS and OKG in Google Ads\footnote{\url{https://ads.google.com/home/}} to conduct a real-world ad campaign.

\paragraph{Experiment Settings.}
For the offline benchmark test, keywords are generated over a time horizon of $T = 5$. At each time step $t$, keywords are adaptively generated by allowing OMS to observe performance feedback. For online A/B test, keywords are generated over a time horizon of $T = 30 Days$, and we generate keywords every 3 days and collect the performance for generation and evaluation.

\subsection{Offline Benchmark Results}
We conducted experiments to compare the OMS with other baselines in the benchmark data. 
The visualization results of all products are in Figure \ref{fig::vis} with normalized numerical values in Table \ref{tab:method_comparison}.
From this table, we can obtain:

\begin{table}
\centering

\scriptsize
\setlength{\tabcolsep}{2.5pt}
\begin{tabular}{@{}lcccccc@{}}
\toprule
\textbf{Method} & \textbf{Clicks}$\uparrow$ & \textbf{CpC}$\downarrow$ &\textbf{\makecell{Search\\ Volume}}$\uparrow$&\textbf{\makecell{Comp\\Score}}$\uparrow$& \textbf{Rouge-1}$\uparrow$ & \textbf{\makecell{BERT\\Score}}$\uparrow$ \\ \midrule
Google KW & 0.70 & 1.00 & 0.62 & 0.73 & 0.16 & 0.41 \\
GPT-4o & 0.74 & 0.95 & 0.79 & 0.77 & 0.18 & 0.54 \\
o3 & 0.74 & 0.95 & 0.59 & 0.77 & 0.18 & 0.54 \\
GPT-4.1 & 0.80 & 0.90 & 0.80 & 0.70 & 0.18 & 0.54 \\
OKG & 0.85 & 0.86 & 0.88 & 0.83 & 0.21 & 0.55 \\
\quad + Many-Shot & 0.89 & 0.88 & 0.82 & 0.72 & 0.20 & 0.56 \\
OMS & \textbf{0.92} & \textbf{0.84} & \textbf{0.90} & \textbf{0.89} & \textbf{0.23} & \textbf{0.62} \\
\bottomrule
\end{tabular}
\vspace{-1em}
\caption{Averaged normalized results of the benchmark. }
\label{tab: numerical results}
\end{table}

\begin{asparaenum}[(I)]
\item The OMS method outperforms baseline methods in all metrics, showing the effectiveness of the proposed method.
The Many-Shot version OKG shows complicated results. It outperforms OKG in some metrics while showing lower performance in other metrics, showing its instability in learning from examples.

\item The naive Google KW, GPT-4o, GPT-4.1, and o3 show a lower performance compared to both OKG and OMS, indicating that general solutions are not yet suitable for a highly customized situation like SSA keyword generation, showing the necessity of developing an agentic framework.

\item The OMS has a higher performance in the Rouge-1 and BERTScore.
This shows that the generated keywords are more related to the product information than keywords generated by other baselines, meaning the generated keywords may attract customers with higher purchase potential.

\end{asparaenum}

\begin{table}
\centering

\scriptsize
\setlength{\tabcolsep}{2.5pt}
\begin{tabular}{@{}lcccccc@{}}
\toprule
\textbf{Method} & \textbf{Clicks}$\uparrow$ & \textbf{CpC}$\downarrow$ &\textbf{\makecell{Search\\ Volume}}$\uparrow$&\textbf{\makecell{Comp\\Score}}$\uparrow$& \textbf{Rouge-1}$\uparrow$ & \textbf{\makecell{BERT\\Score}}$\uparrow$ \\ \midrule
OMS  &  592     & 0.16  &\textbf{2866}&  \textbf{92.7} & \textbf{0.25}      &  \textbf{0.63}         \\
\midrule
\multicolumn{6}{c}{\textbf{Agentic Clustering-Ranking}} \\
\midrule
-w/o Intent &    420   &  0.17  & 1166& 88.7&    0.17     &  0.58         \\
-w/o TOPSIS &    328   &  0.28  & 866& 84.3&    0.14     &  0.56         \\
\midrule
\multicolumn{6}{c}{\textbf{Generation and Multi-turn Reflection}} \\
\midrule
-w/o Reflection   &   431     &  0.18  &1923&90.3&    0.16     &   0.56        \\ 
-w/o LAR &  341      & 0.19  &1324&90.1  &   0.22      &  0.60         \\
\midrule
\multicolumn{6}{c}{\textbf{Single Objective Optimization}} \\
\midrule
OMS Click  &  \textbf{778}  &0.21&  1423   &  91.7   &   0.20 &  0.59    \\
OKG Click &  638  &0.22&  1823   &  90.7   &   0.20  &  0.62 \\ 
OMS CpC  &   451     &  \textbf{0.14} &1889 & 89.6  &   0.21      &     0.60      \\
OKG  CpC & 348& 0.16& 1903 & 90.1 &0.22& 0.58\\
\bottomrule
\end{tabular}
\vspace{-1em}
\caption{Ablation Study over Alpha Camera product. Intent means Intention Analysis.  LAR means the LLM Assisted Re-Clustering.}
\label{tab:ablation}
\end{table}

\subsection{Effects of Shots and Rounds}
 
We analyzed how the number of few-shot examples and generation rounds ($T$) affect performance, demonstrating OMS's ability to utilize examples and maintain keyword quality over time.

\paragraph{Number of Shots}  
(I) For Many-shot OKG, performance is unstable and drops when using over 100 examples, likely due to noise in individual keywords that the LLM cannot effectively learn from. 
(II) In contrast, OMS improves as more examples are added. 
While its performance slightly drops when the example pool is small, as meaningful clusters are not been constructed yet.
After the accumulation of example pools, we see the performance increases as clustered keyword examples with intention analysis could filter noise and provide more informative generation direction.
\paragraph{Number of Rounds}  
(I) Both methods are augmented with 80 examples, and both methods improve over multiple rounds, but OMS improves faster and reaches a higher performance ceiling. 
(II) Multishot OKG suffers from semantic drift as it does not evaluate keyword quality, reflected by declining BERTScore and Rouge-1 with the increase of rounds.
However, OMS maintains a strong relation to the product across rounds due to the self-reflection over keyword quality.

\begin{table*}[t]
  \scriptsize
  \renewcommand{\arraystretch}{2}
  \setlength{\tabcolsep}{2pt}
  \begin{tabularx}{\textwidth}{>{\raggedright\arraybackslash}p{0.34\textwidth} X}
    \toprule
    \textbf{Product Information} & \textbf{Generated Keywords}\\
    \midrule
    \multirow{5}{=}{%
  \begin{minipage}[t]{0.34\textwidth}
\adjustbox{bgcolor=violet!20, padding=0ex 0ex}{Sony \ensuremath{\alpha}-series mirrorless cameras}
    use
    \adjustbox{bgcolor=brown!20 ,padding=0ex 0ex}{full- frame } \& \adjustbox{bgcolor=brown!20 ,padding=0ex 0ex}{APS-C bodies}
    with breakthrough
    \adjustbox{bgcolor=red!20   ,padding=0ex 0ex}{stacked global shutter} sensors.
    The flagship \ensuremath{\alpha}9 III captures action at
    \adjustbox{bgcolor=cyan!20  ,padding=0ex 0ex}{1/80 000 s} and up to
    \adjustbox{bgcolor=blue!15  ,padding=0ex 0ex}{120 fps blackout-free bursts},
    freezing decisive moments without distortion.
    A powerful BIONZ XR engine and
    \adjustbox{bgcolor=green!15 ,padding=0ex 0ex}{AI processing unit}
    enable \adjustbox{bgcolor=green!15 ,padding=0ex 0ex}{real-time tracking} AF that
    recognizes people, animals, birds, insects, and vehicles with skeletal awareness.
    The diverse lineup spans the high-resolution
    \adjustbox{bgcolor=magenta!15,padding=0ex 0ex}{61 MP \ensuremath{\alpha}7CR},
    wildlife-focused \ensuremath{\alpha}7R V, and the
    \adjustbox{bgcolor=yellow!20 ,padding=0ex 0ex}{low-light \ensuremath{\alpha}7S III},
    plus the APS-C
    \adjustbox{bgcolor=lime!15  ,padding=0ex 0ex}{\ensuremath{\alpha}6700 with enhanced eye-AF}.
    Creators can choose from
    \adjustbox{bgcolor=olive!15 ,padding=0ex 0ex}{70+ E-mount lenses}—including the lightweight
    FE 24-50 mm F 2.8 G—for versatile, high-quality imaging.
    Workflow is streamlined via
    \adjustbox{bgcolor=olive!20 ,padding=0ex 0ex}{Creators’ Cloud integration},
    allowing direct uploads to Google Drive and Adobe Photoshop, while select bodies feature
    \adjustbox{bgcolor=teal!20  ,padding=0ex 0ex}{C2PA authenticity signatures}
    to guarantee tamper-proof journalism.
    Together, the \ensuremath{\alpha}-series unleashes high-speed capture and seamless AI-era production for professionals and creators alike.
  \end{minipage}%
} &
\textbf{GPT-4:}
\adjustbox{bgcolor=violet!20, padding=0ex 0ex}{Sony \ensuremath{\alpha} camera},
\adjustbox{bgcolor=brown!20,  padding=0ex 0ex}{\ensuremath{\alpha}9 III mirrorless},
\adjustbox{bgcolor=brown!20,  padding=0ex 0ex}{\ensuremath{\alpha}7 IV full-frame},
\adjustbox{bgcolor=green!15,  padding=0ex 0ex}{AI tracking},
\adjustbox{bgcolor=cyan!20,   padding=0ex 0ex}{High-speed shutter camera},
BIONZ XR image processing,
Sports photography camera, High quality Sony Camera\\[0.3em]
     &
    \textbf{GPT-4.1:}\adjustbox{bgcolor=green!15,  padding=0ex 0ex}{Animal photography AI focus},
Bird and landscape photography,
Video production camera,
\adjustbox{bgcolor=green!15,  padding=0ex 0ex}{AI-powered camera},
\adjustbox{bgcolor=green!15,  padding=0ex 0ex}{AI camera with automatic recognition},
\adjustbox{bgcolor=olive!20,  padding=0ex 0ex}{Photo cloud integration},
Smooth editing camera,
\adjustbox{bgcolor=olive!15,  padding=0ex 0ex}{G Master lens}\\[0.3em]
     &
    \textbf{o3:} \adjustbox{bgcolor=violet!20, padding=0ex 0ex}{Sony \ensuremath{\alpha}9 III},
\adjustbox{bgcolor=blue!15,   padding=0ex 0ex}{Blackout-free shooting},
\adjustbox{bgcolor=magenta!15,padding=0ex 0ex}{\ensuremath{\alpha}
7 CR 61 MP},
\ensuremath{\alpha}
7 R V 61 MP,
High-resolution mirrorless,
\adjustbox{bgcolor=yellow!20, padding=0ex 0ex}{\ensuremath{\alpha}
7 S III for video},
\adjustbox{bgcolor=yellow!20, padding=0ex 0ex}{\ensuremath{\alpha}
7 S III high sensitivity},
\adjustbox{bgcolor=yellow!20, padding=0ex 0ex}{Low-light shooting camera},
\adjustbox{bgcolor=olive!15,  padding=0ex 0ex}{Sony FE 24-50 mm F2.8},
\adjustbox{bgcolor=olive!15,  padding=0ex 0ex}{Lightweight E-mount lens}\\[0.3em]
     &
\textbf{OKG:} 	\adjustbox{bgcolor=violet!20, padding=0ex 0ex}{Sony flagship \ensuremath{\alpha}
 camera},
\adjustbox{bgcolor=blue!15,   padding=0ex 0ex}{High-speed shooting camera},
\adjustbox{bgcolor=magenta!15,padding=0ex 0ex}{High-resolution camera},
\adjustbox{bgcolor=magenta!15,padding=0ex 0ex}{High-res landscape camera},
\adjustbox{bgcolor=cyan!20,   padding=0ex 0ex}{\ensuremath{\alpha}
9 III high-speed shutter},
Instant capture camera,
\adjustbox{bgcolor=green!15,  padding=0ex 0ex}{Real-time tracking camera},
Compact lightweight camera,
Noise-reduction capable camera,
\adjustbox{bgcolor=olive!20,  padding=0ex 0ex}{Automatic photo upload camera},
\adjustbox{bgcolor=teal!20,   padding=0ex 0ex}{Authenticity-signature camera}\\[0.3em]
    &
    \textbf{OMS:}\adjustbox{bgcolor=cyan!20,   padding=0ex 0ex}{High-speed shutter camera},
\adjustbox{bgcolor=blue!15,   padding=0ex 0ex}{120 fps \ensuremath{\alpha}
9 III},
\adjustbox{bgcolor=blue!15,   padding=0ex 0ex}{Mirrorless sports camera},
\adjustbox{bgcolor=green!15,  padding=0ex 0ex}{Real-time tracking AF},
\adjustbox{bgcolor=green!15,  padding=0ex 0ex}{Eye AF camera},
\adjustbox{bgcolor=magenta!15,padding=0ex 0ex}{61 MP \ensuremath{\alpha}
7 CR},
\adjustbox{bgcolor=yellow!20, padding=0ex 0ex}{Low-noise video recording},
\adjustbox{bgcolor=olive!20,  padding=0ex 0ex}{Creators’ Cloud integration},
\adjustbox{bgcolor=olive!20,  padding=0ex 0ex}{Direct upload to Photoshop},
\adjustbox{bgcolor=teal!20,   padding=0ex 0ex}{Authenticity-signature support}\\
    \bottomrule
  \end{tabularx}
  \caption{Case study for the Sony Neural Network Console Product with core features labeled with different colors.}
  \label{tab:aurora_keywords}
\end{table*}




\subsection{Abalation Study}

We conducted an ablation study to assess the importance of each OMS component with results in Table~\ref{tab:ablation}. We can obtain that:

\begin{asparaenum}[(I)]

\item \textbf{Agentic Clustering-Ranking:}  
Removing either intent analysis or the TOPSIS score reduces performance. 
Particularly, removing TOPSIS by prompting with raw, unprocessed metric values significantly degrades results.
Without a unified performance objective, the LLM struggles to make decisions, especially under multiple metrics, and even fails to optimize a single metric among them.

\item \textbf{Multi-Turn Reflection:}  
Disabling the reflection step results in lower Rouge-1 and BERTScore, indicating weaker alignment with product content. Removing LOC-based clustering and relying only on embeddings also hurts performance, as short, similar keywords form low-quality clusters that mislead the generation process.

\item \textbf{Single-Objective Optimization:}  
Optimizing only one metric can improve that metric, but overall performance drops. OMS still outperforms OKG even under single-objective settings.
\end{asparaenum}

\begin{table}[t]\scriptsize
\centering
\setlength{\tabcolsep}{2.5pt}
\begin{tabular}{@{}lcccccc@{}}
\toprule
\textbf{Method}   & \textbf{Cove}$\uparrow$  & \textbf{Rele}$\uparrow$  & \textbf{Spec}$\uparrow$  & \textbf{Redund}$\downarrow$ &  \textbf{Align}$\uparrow$&  \textbf{Overall}$\uparrow$ \\ 
\midrule
GPT-4o &   3.2   &  3.0        &    3.2        &     3.8     & 3.0&2.6    \\
GPT-4.1   &   3.2       &    3.1        &       3.2   & 3.8&3.0&  2.6  \\
o3        &     3.1     &        2.9     &     3.4      & 3.3  & 2.8&  3.1  \\
OKG    &    3.2   &   3.4       &     3.4       &   3.3       & 3.3   &3.5 \\
OMS         &  \textbf{3.5}        &   \textbf{3.8}         & \textbf{3.9} & \textbf{3.0}    &\textbf{3.9}&\textbf{3.7} \\ \bottomrule
\end{tabular}
\caption{Human Preference Ranking for Generated Keywords. Cove, Rele, Spec, Redund, Align means Coverage, Relevance, Speciality, Redundancy, Alignment.}
\vspace{-1em}  
\label{tab:my-table}
\end{table}

\subsection{Human Preference and Case Study}

To evaluate human preference, we anonymized the keywords generated by all methods and asked three professional annotators to rate them (1–5) on coverage, relevance, specialty, redundancy, alignment with user search behavior, and overall quality across five products. 
Results are shown in Table~\ref{tab:my-table}, with a case study in Table~\ref{tab:aurora_keywords}.

\begin{asparaenum}[(I)]

\item \textbf{Human Preference:}  
OMS receives the highest scores across all metrics. GPT-4o and GPT-4.1 often generate redundant or generic keywords. 
The o3 model generates more specialized keywords but lacks alignment with real search behavior. 
OKG performs better than GPT models but was still outperformed by OMS, especially in relevance, alignment, and specialty. 
OMS also slightly leads in coverage, capturing more core product features.

\item \textbf{Case Study:}  
OMS covers the most product features among all methods. 
In contrast, o3 over-focuses on niche series names, GPT-4o/4.1 produces overly general (High quality Sony camera) or off-target keywords (Smooth editing), and OKG had similar issues.
OMS generates keywords that were both relevant and specific, balancing product detail and user intent effectively.
\end{asparaenum}

\begin{table}[ht]\scriptsize
\centering
\renewcommand{\arraystretch}{1.2}
\setlength{\tabcolsep}{10pt}
\begin{tabular}{lrrr}
\toprule
\textbf{Metric} & \textbf{OMS} & \textbf{OKG} & \textbf{Relative Gain} \\
\midrule
\multicolumn{4}{c}{\textbf{Performance Related Metrics (Higher is Better)}$\uparrow$} \\
\midrule
Conversion & 33 & 29 & +13.8\% \\
Clicks & 724 & 713 & +1.5\% \\
Impression & 11,586 & 11,422 & +1.4\% \\
CTR & 6.25\% & 6.24\% & +0.2\% \\
C.V Rate & 4.56\% & 4.07\% & +12.0\% \\
\midrule
\multicolumn{4}{c}{\textbf{Cost Related Metrics (Lower is Better)}$\downarrow$} \\
\midrule
CPA & ¥2,805 & ¥3,192 & -12.1\% \\
CPC & ¥128 & ¥130 & -1.5\% \\
Cost & ¥92,569 & ¥92,575 & -0.01\% \\
\bottomrule
\end{tabular}
\vspace{-1em}
\caption{A/B Test Performance Comparison (OMS vs. OKG). CTR means Click Through Rate, indicating the chance of clicking the campaign advertisement. C.V Rate means Click-to-Conversion rate, indicating the chance of conversion after clicking. CPA means the cost per conversion. The cost unit is Japanese Yen.}
\label{tab:abtest_gain}
\end{table}
\subsection{Online A/B Test on Google Ads }

We deployed OMS and OKG in a real Google Ads campaign using an A/B test with the same budget for a campaign period, with results in Table~\ref{tab:abtest_gain}. \footnote{Details of the A/B test are in the Appendix Sec \ref{Abtest}.}

\begin{asparaenum}[(I)]

\item \textbf{Performance Metrics:}  
OMS outperforms OKG across all performance metrics, especially in Conversion and C.V Rate, which are key indicators in real-world advertising, indicating that keywords from OMS are more attractive and focused.

\item \textbf{Cost Efficiency:}  
Under the same budget, OMS achieves lower CPA  and CpC, meaning it obtained more clicks and conversions at a lower cost. 
In SSA bidding, even a slightly higher bid from other advertisers can dominate other keywords in SSA results. 
However, OMS still achieves better results with lower cost, highlighting its real-world efficiency and effectiveness.
\end{asparaenum}

\section{Conclusion}
In this study, we proposed \textbf{OMS}—a SSA keyword generation framework that is \textit{On-the-fly} (requires no offline training, adapts to real-time performance), \textit{Multi-objective} (optimizes multiple metrics with dynamic prompting), and \textit{Self-reflective} (agentically evaluates keyword quality). 
Experiments show that OMS outperforms prior methods across diverse product benchmarks, with stable performance scaling in both shot count and generation rounds. 
Ablation study analyzed the importance of each component.
Human preference and online A/B test confirm its real-world effectiveness.


\section{Limitation}
While we would like to implement methods that are trained on the query-keyword pair to generate keywords, and compare the proposed method with them, most of those datasets are not released, as the query information in the SSA platform is the in-house data of each platform, which links with personal privacy.
Previous methods did not release such data, and thus, we are not able to train such a model.
However, such methods are provided as a tool or API, like the Google Keyword Planner Service or Bing Keyword Planner Service.
Though the technical details of such a keyword planner service are not released and provided in its documentation,  we treat this method as the representative of traditional methods for SSA keyword generation that require query-keyword pair generation.

While we conducted an online A/B test to compare the OKG and OMS, we were not able to extend it into an A/B/n testing to compare multiple methods in a real-world campaign due to budget limitations.
However, as OMS consistently outperforms other baselines both in the benchmark, which covers products of different popularity, and in the online A/B tests, we think OMS could outperform other baseline methods in most situations.

\section{Ethical Considerations}
For the benchmark we used, we have ensured the usage aligns with the data license and its intended usage.
The collected keywords performance during an A/B test does not contain any personal information, as it is intended for a product campaign set by the company, in which we are only able to see the performance and are not able to see who contributed to a certain conversion or click.
Therefore, this research work is not concerned with ethical issues.

For the usage of AI tools, we used ChatGPT for polishing the writing of this paper.
The usage of AI tools is not beyond this scope.






\bibliography{custom}

\appendix
\newpage
\onecolumn
\section{Additional Experimental Explanation}
\label{sec:appendix}

\subsection{Experiment Setting}

The experiment is run 5 times with different random seeds, which means we sample a different number of keywords for the multi-shot analysis in the benchmark results.
The random seeds used are [42, 123, 456, 891, 777].
The temperature for the GPT-4o backbone is set as 0.
The embedding model that is used to create the cluster is bert-base-cased.
The API version of the used OpenAI models is based on the most recent available API at the time of writing.

The keyword generator and intention analysis is a GPT-4o model.
The evaluator that is used to analyze the keyword quality is an o3-mini-high model.

For non-agentic models that are not equipped with tools, we first run the OMS the retrieve information and save this information as the product information.
This product information will be used by other models to serve as the product information.

We randomly sample the 800 keywords as the evaluation data, and the left 200 keywords with their performance are used for the OKG Many-shot or the OMS. 

Based on the log status from the Langsmith output, each generation took around 0.2-0.5\$ based on the number of iterations.
\begin{algorithm}[!ht]
  \caption{Iterative Keyword Generation with Constraint Feedback}
  \textbf{Input:} initial product query $q_0$, maximum iterations $T$, desired keyword budget $B$,\
  \hspace*{4.1em} rejected keyword set $\mathcal{R}$, deployed history set $\mathcal{H}$,\
  \hspace*{4.1em} search--volume threshold $\tau$, category rejection threshold $\theta$\\
  \textbf{Output:} validated keyword set $\mathcal{K}^\star$

  \begin{algorithmic}[1]
    \State $\textit{info} \gets \Call{Search}{q_0}$ \Comment{Search Tool retrieves initial product information}
    \State $\mathcal{V} \gets \emptyset$ \Comment{temporary store for low--volume keywords}
    \For{$t = 1$ to $T$}
        \If{$\Call{IsEnough}{\textit{info}}$}
            \State $\mathcal{K} \gets \Call{Generate}{\textit{info}}$ \Comment{LLM drafts keyword set}
            \State $\mathcal{K}_v \gets \emptyset$ \Comment{validated keywords}
            \ForAll{$k \in \mathcal{K}$}
                \If{$k \in \mathcal{R}$} \Comment{\textsf{Reject Reflection}}
                    \State \textbf{continue}
                \ElsIf{$k \in \mathcal{H}$} \Comment{\textsf{Repeated Keyword Filter}}
                    \State \textbf{continue}
                \ElsIf{$\Call{SearchVolume}{k} < \tau$} \Comment{\textsf{Search Volume Validator}}
                    \State $\mathcal{V} \gets \mathcal{V} \cup \{k\}$
                \Else
                    \State $\mathcal{K}_v \gets \mathcal{K}_v \cup \{k\}$
                \EndIf
            \EndFor
            \If{$|\mathcal{K}_v| \ge B$}
                \State \Return $\mathcal{K}^\star \gets \mathcal{K}_v$
            \EndIf
            \State $\mathcal{P} \gets \Call{LexicalAnalysis}{\mathcal{V}}$ \Comment{common prefixes/suffixes}
            \State $\Call{UpdateGenerator}{\mathcal{P}}$ \Comment{adapt generation constraints}
            \State $\mathcal{C} \gets \Call{CategoryReject}{\mathcal{V}, \theta}$
            \State $\Call{UpdateCategories}{\mathcal{C}}$ \Comment{drop high--reject clusters}
        \Else
            \State $q \gets \Call{NextQuery}{\textit{info}}$
            \State $\textit{info} \gets \textit{info} \cup \Call{Search}{q}$
        \EndIf
    \EndFor
    \State \Return $\mathcal{K}^\star \gets \mathcal{K}_v$ \Comment{May be empty if budget unmet}
  \end{algorithmic}
\end{algorithm}

\subsubsection{A/B Test Setting}
\label{Abtest}
For the product for which we conducted an actual campaign to perform the A/B test, they are both deployed in the United States.
The campaign lasted for 30 days from 2025 March 20, 2025, to April 20.
The daily budget available for both systems is set at 4,000 yen.
For every generation round, the newly generated keywords are deployed on Tuesday and Friday.
This means the performance observation is collected every 4 or 3 days.
We read the accumulated performance table from the API provided by the SSA platform.
The newly generated keywords will be used to replace keywords that did not receive performance or whose performance did not meet our standards, as Google Ads only allows 50 keywords to be deployed at the same time.

\paragraph{Tool Signatures.}
\begin{align*}
  \textsf{Search} &: \Sigma^* \to \mathcal{I},\\
  \textsf{Generate} &: \mathcal{I} \to 2^{\mathcal{K}},\\
  \textsf{RejectReflection} &: \mathcal{K} \to \{0,1\},\\
  \textsf{RepeatedFilter} &: \mathcal{K} \to \{0,1\},\\
  \textsf{SearchVolume} &: \mathcal{K} \to \mathbb{N},\\
  \textsf{LexicalAnalysis} &: 2^{\mathcal{K}} \to 2^{\mathcal{P}},\\
  \textsf{CategoryReject} &: (2^{\mathcal{K}}, \mathbb{R}) \to 2^{\mathcal{C}}.
\end{align*}
\subsection{Tools Introduction}
\label{tools}
\paragraph{Reject Reflextion} This tool will load keywords with poor performance and analyze why those keywords are bad.
The Keyword Generator receives the analysis result to help it generate better keywords.

\paragraph{Search Tool} This tool is based on SerpAPI \footnote{\url{https://serpapi.com/}} which provides Google search results for a query.
For a given product, the product information is made by manually crafted information (if given) and retrieved information through automated searching.
The keyword generator will first formalize a query about the product itself.
Then, the keyword generator will decide whether the information retrieved is enough.
If not, the Keyword Generator will formulate several queries related to the product and retrieve related information.
This process is executed until the Keyword Generator thinks it has enough information for the keyword generation.

\paragraph{Rejected Keyword Filter} For the deployed keywords, some keywords are above our budget after deployment.
For example, a keyword with a high click-per-cost is saved into a local list, and we do not want the keyword generator to generate those keywords.
The generated keywords will be passed to the Rejected Keyword Filter tool. 
This tool will check whether the generated keywords contain a rejected keyword and flag keywords that do.
The Keyword Generator is asked to regenerate this keyword until all keywords are non-rejected.

\paragraph{Repeated Keyword Filter} As we have already provided and deployed keywords in the multi-shot, since the performance of those keywords is already validated, we do not want the Keyword Generator to generate those keywords again.
Therefore, like the Rejected Keyword Filter, this tool will flag keywords that are in the deployed history keyword.
The Keyword Generator is asked to regenerate those flagged repeated keywords.

\paragraph{Search Volume Check} When we deploy a keyword, the SSA platform will perform a simple check to see if a keyword has been searched a certain number of times in the past several months.
If the generated keyword does not meet this requirement, the SSA platform will automatically disable it.
Therefore, in order to make all generated keywords valid keywords. 
We have to call the API provided by the SSA platform to check the search volume of those generated keywords.

\paragraph{Keyword Lexical Analysis} For those rejected keywords with either low performance or identified with low search volume, we will analyze their common n-gram patterns and their common prefix and suffix.
If the count of such prefixes and suffixes is above a certain threshold, we will append those lexical features to the generation rules of the keyword generator.
This makes sure the LLM keyword generator will not get stuck in an endless loop of regenerating keywords in order to pass the search volume validator.

\paragraph{Category Analysis} As the keyword generator generates keywords according to the categories analyzed by the Keyword Generator agent.
However, it is possible that if the category itself is bad, e.g, the category is too niche, then all generated keywords following this category will not pass the final search volume check, no matter how many times of regeneration.
To avoid this situation, we calculate the percentage of rejected keywords for a category across multiple times of generations, then we can identify those categories that have a high reject rate for their keywords.
If such a category is identified, the whole category is rejected, and the LLM agent is asked to generate a whole new category.

\section{Example Workflow}
\begin{figure}[htbp] 
\centering
\includegraphics[width=1\columnwidth]{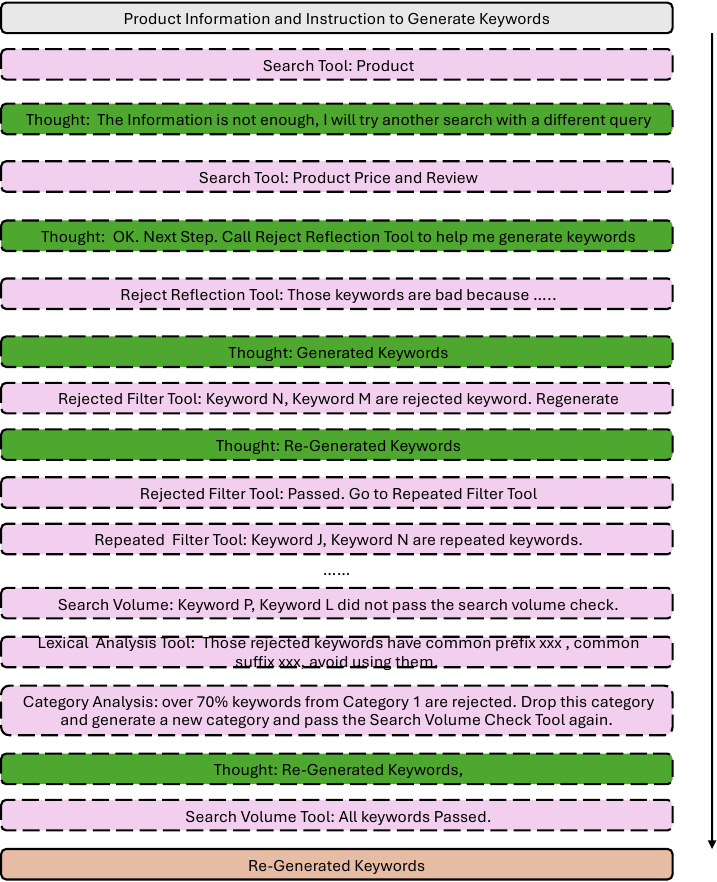} 
\caption{An example workflow for keyword generation.}
\label{fig:flow}
\end{figure}

We present an example workflow of the Tool-Return Guided Generation workflow in the Figure \ref{fig:flow}.

\section{Annotator Description for Human Preference over Generated Keywords}

We asked annotators (they are recruited within the company organization, thus no payment is provided) to evaluate advertisement keywords generated by five anonymized systems (Systems 1 to 5). 
We have obtained their consent to use their evaluation scores.
The evaluation was conducted based on the following six criteria. Each criterion is rated on a scale from 0 to 5, where a higher score indicates better performance (except for Redundancy, where lower is better).

\begin{itemize}
    \item \textbf{Coverage}: How well the keyword set covers the product features described in the product information text.
    \item \textbf{Relevance}: Whether the keyword set avoids including irrelevant or weakly related terms.
    \item \textbf{Specificity}: How specifically the keywords reflect the core characteristics of the product.
    \item \textbf{Redundancy}: Whether there are redundant or similar keywords that could be removed (a lower score indicates better quality).
    \item \textbf{Search User Behavior Alignment}: To what extent the keywords align with the search behavior of users looking for similar products.
    \item \textbf{Overall Quality}: The overall evaluation of the keyword set, considering all the criteria above.
\end{itemize}

The following are the examples we provided to the human preference evaluation task for annotators.
\paragraph{Example 1:}
\textbf{Product Information:}
The handmade dorayaki from “Kyoto Marushin” features chunky sweet bean paste carefully cooked from Hokkaido-grown azuki beans, sandwiched in fluffy, specially made pancakes. Free from preservatives and artificial coloring, it offers a natural sweetness and the original flavor of the ingredients. Ideal for both gifts and personal enjoyment.

\begin{itemize}
\item \textbf{Keywords with high specificity but low relevance:} Azuki sweets, Azuki dessert
\item \textbf{Keywords aligned with user search behavior but low in specificity:} Kyoto travel sweets
\item \textbf{Keywords with low coverage and relevance:} Dorayaki recipe
\item \textbf{Good keyword examples:} Kyoto dorayaki, Kyoto wagashi (traditional sweets), Handmade dorayaki, Additive-free dorayaki
\end{itemize}

\paragraph{Example 2:}
\textbf{Product Information:}
The SHARP Plasmacluster 7000 air purifier supports rooms up to 13 tatami mats in size, effectively removing pollen, dust, and PM2.5 particles. Its proprietary Plasmacluster technology also suppresses airborne viruses and mold. Equipped with a quiet mode, it is ideal for nighttime use and bedrooms. The long-life filter is easy to maintain.

\begin{itemize}
\item \textbf{Keywords aligned with user search behavior but low in coverage or specificity:} High-performance filter, Popular air purifier
\item \textbf{Keywords with high specificity but low coverage:} Plasmacluster
\item \textbf{Criteria for a good keyword set:} Covers key product features (suitable for 13-tatami rooms, pollen/PM2.5 removal, quiet mode, long-life filter), avoids redundancy, and aligns well with user search behavior.
\end{itemize}
\subsection{Human Preference Alignment}
We have three annotators for the agreement between the annotators for the generated keywords.
We anonymized three annotators with A1, A2, and A3.
We show the inner-annotator agreement between annotators in the Table \ref{tab:annotator_agreement}.
The results show that all annotators reach similar judgments regarding the keyword quality, supporting the validity of the experimental results.

\begin{table}[htbp]
\centering
\small

\begin{tabular}{lccc}
\toprule
\textbf{Metric} & \textbf{A1-A2} & \textbf{A1-A3} & \textbf{A2-A3} \\
\midrule
Coverage & 0.607 & 0.796 & 0.626 \\
Relevance & 0.604 & 0.447 & 0.620 \\
Specificity & 0.650 & 0.551 & 0.623 \\
Redundancy & 0.640 & 0.596 & 0.579 \\
Search Behavior & 0.520 & 0.695 & 0.687 \\
Overall Quality & 0.590 & 0.829 & 0.625 \\
\midrule
\textbf{Average} & \textbf{0.601} & \textbf{0.652} & \textbf{0.627} \\
\bottomrule
\end{tabular}
\caption{Inter-annotator Agreement (Kendall's Tau Correlation)}
\label{tab:annotator_agreement}
\end{table}

\subsection{More Generated Keywords Examples}
\paragraph{Example:}
\textbf{Product Information:} \\
Sony's Prediction One is a no-code AutoML tool that allows users without machine learning expertise to build high-accuracy models in a few clicks. Developed by Sony Network Communications, it automates everything from algorithm selection to hyperparameter tuning. It supports business use cases like demand forecasting and churn prediction, and provides APIs for integration into existing systems. A GUI-based interface and team collaboration features enable easy adoption across industries, addressing the AI talent shortage.

\begin{itemize}
    \item o3: AutoML tool, No-code machine learning, High-accuracy model automation, Demand forecasting software, Churn prevention analysis, Failure prediction, Prediction One pricing, Prediction One deployment, Prediction One enterprise plan, AutoML comparison, AutoML tool comparison, Alternative to DataRobot
    \item OKG: No-code machine learning, AutoML tool, Sony Prediction One, demand forecasting software, sales efficiency AI, DX promotion tool, data analysis software, automated prediction model, API prediction integration, AI tool for small businesses, DX solution for small businesses, business automation software
    \item OMS: No-code machine learning, AutoML tool, AI prediction model, demand forecasting system, DX promotion tool, inventory optimization AI, Sony Prediction One, Sony AutoML, Sony AI tool, AI services for businesses, cloud AI implementation, AI tool with support
    \item GPT-4o: High-accuracy model auto-construction, Sony AutoML, evidence-based prediction machine learning, one-click machine learning, Prediction One AI system, multi-industry optimization AI, one-click AI development
    \item GPT-4.1: Sony Prediction One ML system, efficiency improvement ML tool, code-free AI, Prediction One business AI service, explainable prediction AI tool, easy AI implementation, GUI-based ML development, cloud AI tool, machine learning service for enterprises, sales analysis optimization tool
\end{itemize}

\paragraph{Example:}
\textbf{Product Information:} \\
Sony's Neural Network Console is a no-code development environment for designing and training neural networks using drag-and-drop. It integrates with a Python API for hybrid GUI + code development and features automatic architecture search to speed up model optimization. Users can visually compare past experiments and export models in ONNX format for deployment on Edge AI devices. It's used in real Sony products like Aibo and Xperia Ear, and supports beginners through tutorials and corporate users with PoC services.

\begin{itemize}
    \item o3: Neural network development software, Neural network design tool, Deep learning GUI, AI model auto generation, Drag-and-drop machine learning, AutoML hyperparameter tuning, Automatic model structure search, Automated machine learning tool, ONNX export tool, Edge AI model deployment, Mobile AI integration
    \item OKG: Easy neural network development, programming-free machine learning, beginner AI tutorial, image classification learning tool, time series prediction introduction, AI PoC solution, DX support service, AI prototype creation, Edge AI model development, ONNX export method, AI smartphone app integration
    \item OMS:Drag-and-drop AI, deep learning development environment, neural network training, Python-integrated AI tool, AI prototyping, AI PoC support, DX promotion service, ONNX export, edge AI device, smartphone AI integration
    \item GPT-4o: No-code AI development, no-code AI tool, edge AI implementation, AI talent shortage solution, AI business proposal, visualized AI tool, Sony AI product tool, AI transformation support
    \item GPT-4.1: Model structure auto-optimization AI tool, AI tool with implementation examples, Python-integrated AI development, free AI tool, AI deployable to edge devices, deep learning development for beginners, neural network GUI development, no-code AI development
\end{itemize}

\paragraph{Example:}
\textbf{Product Information:} \\
Sony's cloud-based attendance management system "AKASHI" supports diverse work styles from telework to flex-time. It simplifies HR operations with intuitive drag-and-drop rule settings, automatic creation of legally required records, and integration with face recognition AI for secure time tracking. The system enables real-time management from anywhere and helps companies reduce overtime through alerts and automated payroll integration. Offered as a monthly subscription, it’s suitable for mid-sized companies aiming to digitize attendance management.

\begin{itemize}
    \item o3: Attendance management system, Cloud-based attendance software, Telework attendance tracking, Flex-time attendance, 36-agreement compliance tool, Paid leave management system, HR compliance software, Attendance-payroll API integration, Automated attendance data, HR system integration
    \item OKG: Cloud-based attendance management, attendance software for small businesses, attendance system with facial recognition, HR efficiency solution, labor management automation, overtime reduction tool, telework attendance management, remote work clock-in system, remote attendance cloud, work style reform attendance tool, labor risk prevention system, digital back office transformation
    \item OMS: Cloud attendance management, attendance management system implementation, attendance management solution, HR and labor management, business automation system, paperless management, telework attendance management, flex time attendance, work style reform system, payroll system integration, expense reimbursement integration, API-compatible attendance management
    \item GPT-4o: Comprehensive attendance management solution, clock-in/out system, automated management creation, zero initial cost attendance system, work style reform, customizable attendance system, all-in-one attendance system, remote work automatic management
    \item GPT-4.1: Sony cloud-based attendance management system, automated attendance management system, attendance system with implementation examples, data visualization, AKASHI attendance management, productivity improvement attendance management, law-compliant auto-updating attendance management, fraud prevention attendance system, remote clock-in system, automatic labor condition detection system
\end{itemize}

\paragraph{Example:}
\textbf{Product Information:} \\
Sony's next-generation BRAVIA series was developed under the concept of "experiencing the world beyond the screen." By leveraging the latest panel technologies—Mini LED and QD-OLED—it delivers dazzling brightness and deep blacks even in bright living rooms. The BRAVIA 8 II adopts third-gen QD-OLED panels with superior peak brightness and wide color gamut, reproducing vibrant colors for HDR content. The BRAVIA 9 features dense local dimming for rich contrast. With "Auto HDR Tone Mapping" and ultra-low latency (~8 ms), it offers smooth gaming. The BRAVIA CAM adjusts picture and sound based on user position, ideal for both family living rooms and work desks.

\begin{itemize}
    \item o3: Sony QD OLED TV, Mini LED BRAVIA comparison, BRAVIA 9 high brightness, BRAVIA game mode low latency, Sony TV for eSports, Game Menu settings, Auto HDR tone mapping, BRAVIA deep blacks, BRAVIA wide color volume, BRAVIA CAM auto adjustment, Sony TV sound optimization
    \item OKG: QD OLED TV, Mini LED high picture quality, HDR image optimization, low latency gaming TV, game menu supported TV, large screen TV for families, automatic picture adjustment BRAVIA CAM, living room optimized TV
    \item OMS: Mini LED TV, QD OLED TV, high brightness HDR TV, low latency gaming TV, TV with crosshair function for gaming, BRAVIA for eSports, BRAVIA CAM auto adjustment, optimal picture quality BRAVIA, TV with automatic viewing distance adjustment, cinema-level black reproduction TV, vivid color TV
    \item GPT-4o: Living room movie watching TV, eSports optimized TV, BRAVIA for family living room, BRAVIA game mode low latency, automatic adjustment TV, Sony TV latest model
    \item GPT-4.1: Auto-optimized setting TV, multi-purpose TV, low latency TV, Sony BRAVIA TV, BRAVIA new series, Mini LED TV, QD OLED TV, high picture quality TV for gaming, home use TV, cinema experience TV
\end{itemize}





\section{Prompts used in Methodology}
\label{app::prompt}

\subsection{LLM\_Intent Prompt}

The following prompt is used to implement the function \( \texttt{LLM\_Intent}(r, C_j) \) defined in Equation~\ref{eq::intent} for intent analysis of each keyword cluster:

\begin{promptbox}[LLM\_Intent Prompt]
Product name \{product\_name\} and the users searched for the following keywords to reach the website related to this product.

The related product information is given in the following: \{product\_information\}

You are given the following clusters of keywords and their clicks:  
Cluster N: Keyword 1: Click, Cost, Conversion, Impression. ...

Analyze the above cluster and its keywords. Especially check the common features of those keywords and explain why the user searched for those keywords.
\end{promptbox}

\subsection{LLM\_Rank Prompt Template}

This prompt is constructed automatically after computing TOPSIS scores for each keyword and cluster. It is used as input to the function \( \texttt{LLM\_Rank}(C_j) \) defined in Equation~\ref{eq::rank}. The structure below is generated programmatically using a for-loop over clusters and their ranked keywords:

\begin{promptbox}[LLM\_Rank Prompt Template]
You are given clusters of keywords with their corresponding TOPSIS scores and rankings.

Each cluster is listed with its average TOPSIS score, and keywords within each cluster are sorted by their individual scores.

Use this information to determine which clusters or keywords should be prioritized for expansion or refinement.

\textbf{Cluster: \{cluster\_name\} (Avg Score: \{cluster\_avg\_score\})}  
1. \{keyword\_1\} — Score: \{score\_1\}  
2. \{keyword\_2\} — Score: \{score\_2\}  
3. \{keyword\_3\} — Score: \{score\_3\}  
...

\textbf{Cluster: \{cluster\_name\} (Avg Score: \{cluster\_avg\_score\})}  
1. \{keyword\_1\} — Score: \{score\_1\}  
2. \{keyword\_2\} — Score: \{score\_2\}  
...

...

Please analyze the clusters and their keyword rankings. Focus on the strongest-performing clusters and keywords for generation, and suggest improvements for weaker ones.
\end{promptbox}

\subsection{LLM\_Reflect Prompt Template}

This prompt is used to implement the function \( \texttt{LLM\_Reflect}(k_i) = \texttt{LLM}(k_i, K_t, r) \) as defined in Equation~\ref{eq::reflect}. The prompt is dynamically constructed using the product description, current generated keywords, and historical evaluations:

\begin{promptbox}[LLM\_Reflect Prompt Template]
You are given the intermediate generated keyword result (formatted as a dictionary with two main keys: \{'Branded'\} and \{'Non-Branded'\}) and the product information. You should evaluate the coherence between each keyword and the product.

For each keyword, give a score from 1 to 5 based on how well it represents the product information. Also, provide a reason for the score and suggest whether the keyword should be kept or replaced.

Scoring guide:  
1: The keyword does not represent the product at all.  
2: The keyword poorly represents the product.  
3: The keyword somewhat represents the product.  
4: The keyword represents the product well.  
5: The keyword perfectly represents the product.

A good keyword should capture key product features and not be overly generic. Consider both semantic relevance and user search behavior.

Provide your output in the following dictionary format:
\{
  "keyword": "\{keyword\}",  
  "score": \{score\},  
  "reason": "\{reason\}",  
  "suggestion": "\{keep/replace\}"  
\}

The generated keywords to evaluate are:  
\{generated\_keywords\}

The product information is:  
\{product\_information\}

Your evaluation history is:  
\{history\_evaluation\}

Only evaluate keywords not already included in the history.
\end{promptbox}

\subsection{LLM\_Assign Prompt Template}

This prompt is used to implement the function \( \texttt{LLM\_Assign}(k_i, \mathcal{C}_{k_i}^{\text{top}}) \) as defined in Equation~\ref{formu::recluter}. The prompt is constructed dynamically using the target keyword, the top-3 nearest clusters (based on embedding similarity), and the product information:

\begin{promptbox}[LLM\_Assign Prompt Template]
You are given a keyword:  
\{keyword\_tobe\_decided\}

The following three clusters are the closest clusters to this keyword based on embedding similarity:

The product information is:  
\{product\_information\}

Cluster 1:  
\{cluster\_1\_keywords\}

Cluster 2:  
\{cluster\_2\_keywords\}

Cluster 3:  
\{cluster\_3\_keywords\}

Based on the product information and the semantic intent of each cluster, decide whether the given keyword should be assigned to one of the above clusters or treated as a new cluster.

Please respond with exactly one of the following options:  
\texttt{Cluster 1}, \texttt{Cluster 2}, \texttt{Cluster 3}, or \texttt{New Cluster}.
\end{promptbox}

\subsection{LLM\_Generate Prompt Template}

This static prompt is given to the agent at initialization to define the overall task. It instructs the agent to generate advertising keywords through multi-step reasoning based on product information, past failures, and user search behavior.

\begin{promptbox}[LLM\_Generate Prompt Template]
You are tasked with generating advertising keywords for \{product\_name\}.

Your keywords must reflect the product’s key features and align with how users typically search online. Avoid technical terms and duplicates from previous keywords.

Before generation, use tools such as \texttt{google\_search} to gather product information and \texttt{reject\_reflection} to analyze failed keywords.

Your final output must be a dictionary-like string with two keys: \texttt{"Branded"} and \texttt{"Non-Branded"}, each containing 10 high-quality keywords.
\end{promptbox}

\end{document}